\documentclass[letterpaper, 10 pt, conference]{ieeeconf}
\IEEEoverridecommandlockouts
\usepackage{cite}
\usepackage{amsmath,amssymb,amsfonts}
\usepackage{algorithmic}
\usepackage{graphicx}
\usepackage{textcomp}
\usepackage{colortbl}
\usepackage{float}
\usepackage{booktabs}
\usepackage[table]{xcolor}
\definecolor{maroon}{rgb}{0.6, 0.87, 0.94}
\definecolor{awesome}{rgb}{0.98, 0.66, 0.68}
\usepackage{multirow}
\usepackage{etoolbox}
\usepackage{setspace}
\usepackage{etoolbox}
\usepackage{multirow}
\usepackage{algorithmic}
\usepackage{algorithm}
\usepackage{threeparttable}

\usepackage{hyperref}
\makeatletter

\title{\LARGE \bf
VLA Model Post-Training via Action-Chunked PPO \\ and Self Behavior Cloning}

\begin{document}

\author{Si-Cheng Wang$^\dagger$, Tian-Yu Xiang$^\dagger$, Xiao-Hu Zhou$^*$, Mei-Jiang Gui, Xiao-Liang Xie, Shi-Qi Liu, \\Shuang-Yi Wang, Ao-Qun Jin, Zeng-Guang Hou
\thanks{This work was supported in part by the National Key Research and Development Program of China under 2023YFC2415100, in part by the National Natural Science Foundation of China under Grant 62222316, Grant 62373351, Grant 82327801, Grant 62073325, Grant 62303463, in part by the Chinese Academy of Sciences Project for Young Scientists in Basic Research under Grant No.YSBR-104 and in part by China Postdoctoral Science Foundation under Grant 2024M763535.}
\thanks{The authors are with State Key Laboratory of Multimodal Artificial Intelligence Systems, Institute of Automation, Chinese Academy of Sciences, and also with School of Artificial Intelligence, University of Chinese Academy of Sciences, Beijing 100049, China. E-mail: $\{$xiangtianyu2021, xiaohu.zhou, zengguang.hou$\}$@ia.ac.cn}
\thanks{$\dagger$ Equally contribution: Si-Cheng Wang, Tian-Yu Xiang, $*$ Corresponding author: Xiao-Hu Zhou}
}

\maketitle
\thispagestyle{empty}
\pagestyle{empty}

\begin{abstract}

Reinforcement learning (RL) is a promising avenue for post-training vision–language–action (VLA) models, but practical deployment is hindered by sparse rewards and unstable training. This work mitigates these challenges by introducing an action chunk based on proximal policy optimization (PPO) with behavior cloning using self-collected demonstrations. Aggregating consecutive actions into chunks improves the temporal consistency of the policy and the density of informative feedback. In addition, an auxiliary behavior cloning loss is applied with a dynamically updated demonstration buffer that continually collects high-quality task trials during training. The relative weight between the action-chunked PPO objective and the self behavior clone auxiliary loss is adapted online to stabilize the post-training process. Experiments on the MetaWorld benchmark indicate improved performance over supervised fine-tuning, achieving a high success rate (0.93) and few steps to success (42.17). These results demonstrate the viability of RL for VLA post-training and help lay the groundwork for downstream VLA applications.

\end{abstract}


\section{Introduction}

Motivated by advances in multimodal vision–language models (VLMs)\cite{zhang2024vision}, recent studies have begun exploring foundation models for robotics\cite{firoozi2023foundation}. A direct extension augments VLMs with an action modality, yielding vision–language–action (VLA) models. By inheriting strong vision-language understanding of VLMs, VLA models shows strong manipulation potentials~\cite{ma2024survey}. Nevertheless, the scarcity of large-scale manipulation datasets and the heterogeneity of robotic embodiments constrain zero-shot performance, making post-training a practical necessity for deployment~\cite{xiang2025parallels}.

A common post-training strategy is to collect demonstrations and apply supervised fine-tuning to learn a mapping from observations to actions. This approach is straightforward and has shown effectiveness in adapting VLA models to downstream tasks. However, obtaining high-quality demonstrations is costly. Although the required demonstration set for post-training is much smaller than that for pre-training, it typically still comprises on the order of several dozen to about one hundred demonstrations~\cite{2024arXiv240512213O, kimopenvla}, and the acquisition process remains labor-intensive and time-consuming.

RL offers a promising way to address the above problem and has been applied in robotics for decades~\cite{kober2013reinforcement}. Unlike supervised fine-tuning, RL adapts a policy to tasks through iterative interaction with the environment under a reward function. However, deploying a VLA model within an RL framework presents two challenges. First, in many manipulation settings the reward signal is sparse, which complicates credit assignment and slows training~\cite{nam2023lift}. Second, the shift from supervised pre-training to RL-based optimization can introduce instability due to the noises of policy-gradient~\cite{hu2024flare}.

To provide dense reward during RL based post-training, prior work incorporates external expert feedback from either humans~\cite{chen2025conrft} or VLMs~\cite{lee2024affordance}. Human feedback is typically more reliable, whereas VLM feedback is automated and avoids annotation labor. In both cases, denser signals can accelerate convergence, but at the cost of additional system overhead (human effort or computational expense). To further stabilize training, a small demonstration set (e.g., $\sim$10 trials) is commonly collected and a supervised behavior-cloning term is added to the RL objective~\cite{hu2024flare}. Combined with standard stabilization techniques~\cite{fujimoto2018addressing, chenrandomized}\cite{chenrandomized}, this hybrid scheme has shown effectiveness. However, because the demonstrations are limited and often suboptimal, the behavior-cloning term can become detrimental once the VLA policy surpasses demonstration performance.

To address the above limitations, an action-chunked RL algorithm based on PPO with self-collected demonstrations is proposed for post-training VLA models. The method aggregates consecutive actions into action chunks, increasing the frequency of reward feedback in PPO. A dynamic demonstration buffer integrates high-quality trajectories generated by the agent for constructing an auxiliary supervised loss. The contributions of this study are summarized as follows:

\begin{figure*}[!tb]
    \centering
      \includegraphics[width=7in]{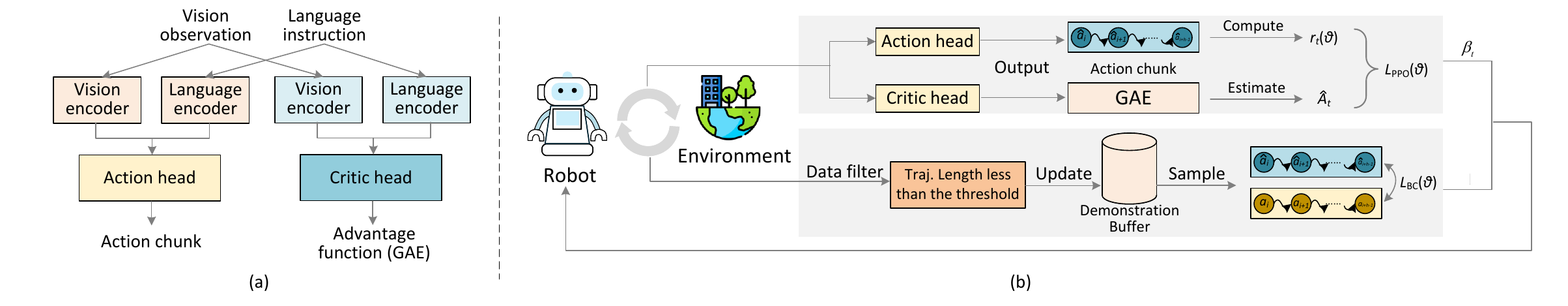}
        \caption{Pipeline of the proposed method. (a) The actor–critic architecture. (b) Post-training under hybrid action-chunked PPO and self-behavior cloning.} 
     \centering
     \label{fig:pipeline}
 \end{figure*}

\begin{itemize}
    \item An action-chunked PPO algorithm is developed for post-training VLA models, increasing the effective density of informative feedback.
    \item A self behavior cloning auxiliary loss is constructed based on the dynamically demonstration buffer that consistently collects high-quality task trials during learning.
    \item Experiments show that the method initialized with only $10$ demonstrations surpasses supervised fine-tuning with $100$ demonstrations in both success rate (0.93 vs.\ 0.89) and steps to success (42.17 vs.\ 65.65).
\end{itemize}

\section{Method}

The proposed algorithm is designed for post-training the VLA model within the online setting, involving an action-chunked PPO and a supervised self behavior cloning term (Fig.~\ref{fig:pipeline}).

\subsection{Action-chunked PPO}

Let $\pi_{\theta}$ denotes the VLA policy with parameters $\theta$. At time step $t$, given the observation $o_t$ and prompt $p_t$, the policy outputs an action chunk of length $h\!\ge\!1$ composed of a sequence of actions ${a}$,
\begin{equation}
\mathbf{a}_{t:t+h-1} \triangleq [a_t, a_{t+1}, \ldots, a_{t+h-1}] = \pi_{\theta}(o_t, p_t).
\end{equation}

The PPO framework is realized based on the actor-critic architecture~\cite{2016arXiv160201783M}. During training, policy updates are constrained by a clipped surrogate to limit the divergence between the behavior and current policies, reducing update-induced instability. Gradient directions are governed by the advantage function $\hat{A}_t$, computed via generalized advantage estimation (GAE)~\cite{2015arXiv150602438S} from observed rewards and the learned value function. The policy is optimized under the PPO objective $\mathcal{L}_{\mathrm{PPO}}$ given below:
\begin{equation}
    \mathcal{L}_{\mathrm{PPO}}(\theta) = \mathbb{E}_{t} \left[ \min\left( r_t(\theta) \hat{A}_t, \text{clip}(r_t(\theta), 1-\epsilon, 1+\epsilon) \hat{A}_t \right) \right]
\end{equation}
where $\text{clip}(x, a, b) = \min(\max(x,a),b)$, $\epsilon$ is a clipping hyperparameter, $r_t$ denotes the likelihood ratio between the current and old policies for the action chunk at time $t$:
\begin{equation}
    r_t(\theta) = \frac{\pi_\theta(\mathbf{a}_{t: t+h-1}|o_{t},p_{t})}  {\pi_{\theta_{\text{old}}}(\mathbf{a}_{t: t+h-1}|o_{t},p_{t})}
\end{equation}
where, $\pi_\theta$ and $\pi_{\theta_{\text{old}}}$ represent the current and old policy, respectively.

The value function (critic head, which is denoted as $V_{\phi}$), parameterized by $\phi$, is optimized using the clipped value loss ($\mathcal{L}_{\text {critic}}$) from PPO to prevent the new policy from deviating too much from the old one:
\begin{equation}
\begin{split}
    \mathcal{L}_{\text {critic}}({\phi}) = & \mathbb{E}_{(o_{t},p_{t}) \sim \pi_{\text {old }}}[ \max \left(\left[R(o_{t},p_{t})-V_{\phi}(o_{t},p_{t})\right]^{2}, \right. \\
&\left. \left[R(o_{t},p_{t})-\operatorname{clip}\left(V_{\phi}(o_{t},p_{t}), V_{\phi_{\text {old }}}(o_{t},p_{t})-\epsilon, \right.\right. \right. \\
&\left.\left. \left.  V_{\phi_{\text {old }}}(o_{t},p_{t})+\epsilon\right)\right]^{2}\right)]
\end{split}
\end{equation}
where the reward $R$ is expressed as:
\begin{equation}
    R\left(o_{t},p_{t}\right) = R_t + \gamma V_{\phi_{\text {old }}}\left(o_{t+1},p_{t+1}\right)
\end{equation}
where $R_t$ represents the immediate reward at time $t$, $\gamma$ is the discount factor.

\subsection{Self Behavior Cloning}

Post-training a VLA policy with PPO can be unstable due to high-variance RL gradients~\cite{hu2024flare}. To mitigate this effect, a supervised self behavior cloning auxiliary objective ($\mathcal{L}_{\mathrm{BC}}$) is employed:
\begin{equation}
    \mathcal{L}_{\mathrm{BC}}(\theta) = \mathbb{E}_{(\mathbf{a}_{t},o_{t},p_{t}) \sim \mathcal{D}_{\text{demo}}} \left[ -\log \pi_\theta(\mathbf{a}_{t: t+h-1}|o_{t},p_{t}) \right]
\end{equation}
where $\mathcal{D}_{\text{demo}}$ is a dynamic demonstration buffer. The buffer is initialized with expert trajectories. During training, when the agent produces new high-quality successful trajectories $x_i$, it will be included into the buffer. This process increases the diversity of demonstrations and can improve data quality, as some agent-generated demonstrations may exceed the quality of the initial expert data.

In implementation, trajectory quality is proxied by trajectory length, under the assumption that faster task completion typically indicates a more effective policy. Let the length of a trajectory $x_i$ be denoted by $L(x_i)$, with smaller values indicating higher quality. Trajectories are retained if they satisfy a length threshold: $L(x_i)\ \le\ \ell_{\text{limit}}$. The demonstration dataset is initialized with a small set of expert trajectories, and the threshold ($\ell_{\text{limit}}$) is set to the longest expert trajectory.

\begin{table*}[htb!] \small
\renewcommand{\arraystretch}{1.2} 
\centering
\caption{Experimental results on MT10 benchmark of MetaWorld environment}
\begin{tabular}{p{2.95cm}|
                p{0.8cm}<{\centering}
                p{0.8cm}<{\centering}
                p{0.8cm}<{\centering}
                p{0.8cm}<{\centering}|
                p{0.8cm}<{\centering}
                p{0.8cm}<{\centering}
                p{0.8cm}<{\centering}
                p{0.8cm}<{\centering}|
                p{0.8cm}<{\centering}
                p{0.8cm}<{\centering}
                p{0.8cm}<{\centering}
                p{0.8cm}<{\centering}}
\toprule
\multirow{2}{*}{\textbf{Task}} & \multicolumn{4}{c|}{\textbf{Acc.}} & \multicolumn{4}{c|}{\textbf{Len.}} & \multicolumn{4}{c|}{\textbf{Avg(l).}} \\
& \textbf{SFT} & \textbf{SFT$_{100}$} & \textbf{PPO} & \textbf{Ours} & \textbf{SFT} & \textbf{SFT$_{100}$} & \textbf{PPO} & \textbf{Ours} & \textbf{SFT} & \textbf{SFT$_{100}$} & \textbf{PPO} & \textbf{Ours} \\
\hline
Reach & 0.29 & 0.32 & 0.78 & \textbf{0.93} & 42.00 & 42.00 & 31.00 & \textbf{26.00} & 32.75 & 35.42 & 29.17 & \textbf{23.75} \\
Push & 0.34 & \textbf{0.96} & 0.00 & 0.77 & 59.00 & 54.00 & - & \textbf{41.00} & 55.42 & 52.58 & - & \textbf{38.08} \\
Pick-Place & 0.34 & 0.83 & 0.00 & \textbf{0.85} & 51.00 & 50.00 & - & \textbf{39.00} & 48.00 & 47.92 & - & \textbf{36.00} \\
Door-Open & 0.99 & \textbf{1.00} & 0.40 & 0.97 & 75.00 & 74.00 & 89.00 & \textbf{62.00} & 73.67 & 73.50 & 86.92 & \textbf{61.25} \\
Drawer-Open & 0.92 & \textbf{1.00} & 0.44 & \textbf{1.00} & 87.00 & 89.00 & 59.00 & \textbf{39.00} & 86.67 & 89.00 & 58.83 & \textbf{37.25} \\
Drawer-Close & \textbf{1.00} & \textbf{1.00} & 0.47 & 0.98 & 78.00 & 77.00 & 36.00 & \textbf{18.00} & 78.00 & 77.00 & 34.92 & \textbf{18.00} \\
Button-Press-Topdown & 0.94 & \textbf{1.00} & 0.00 & 0.95 & 60.00 & 59.00 & - & \textbf{54.00} & 59.17 & 58.92 & - & \textbf{52.50} \\
Peg-Insert-Side & 0.20 & 0.76 & 0.00 & \textbf{0.84} & 87.00 & 76.00 & - & \textbf{64.00} & 79.67 & 66.75 & - & \textbf{58.91} \\
Window-Open & 0.93 & \textbf{1.00} & 0.28 & 0.99 & 80.00 & 80.00 & 49.00 & \textbf{47.00} & 76.83 & 78.42 & 44.50 & \textbf{44.08} \\
Window-Close & \textbf{1.00} & \textbf{1.00} & 0.00 & 0.99 & 76.00 & 77.00 & - & \textbf{53.00} & 76.00 & 77.00 & - & \textbf{51.83} \\
\hline
Average & 0.70 & 0.89 & 0.24 & \textbf{0.93} & 69.50 & 67.80 & - & \textbf{44.30} & 66.62 & 65.65 & - & \textbf{42.17} \\
\bottomrule
\end{tabular}
\label{tb1: results}
\end{table*}

\subsection{Online Post-training}

The online post-training objective for the VLA model is a weighted combination of the PPO loss $\mathcal{L}_{\mathrm{PPO}}$ and the supervised behavior clone loss $\mathcal{L}_{\mathrm{BC}}$:
\begin{equation}
    \mathcal{L}_{\mathrm{online}}(\theta) = \beta_t \mathcal{L}_{\mathrm{PPO}}(\theta) + \mathcal{L}_{\mathrm{BC}}(\theta)
\end{equation}
where $\beta_t$ adjusts the relative weight of PPO to BC at training step $t$. In early training, the critic may be poorly aligned with the actor, and applying online RL can degrade the policy due to biased or high-variance value estimates. To mitigate this issue, a progressive schedule is used:
\begin{equation}
    \beta_t = \tanh\left( \frac{t}{T_{\text{warmup}}} \right)
\end{equation}
where $T_{\text{warmup}}$ denotes the number of warm-up steps. At the beginning ($t=0$), the optimization is driven solely by $\mathcal{L}_{\mathrm{BC}}$. As training progresses, $\beta_t$ increases monotonically, gradually emphasizing $\mathcal{L}_{\mathrm{PPO}}$ and yielding a smooth transition from supervised imitation to reinforcement learning.

\section{Experiments and Results}

\subsection{Implementation Details}

\subsubsection{Experimental Environment}

Experiments are conducted on MetaWorld~\cite{2019arXiv191010897Y}, using 10 single-task environments from MT10 with sparse rewards. The demonstration dataset is collected with the rule-based policy provided with the environment, each trajectory capped at 200 steps. Data are recorded from a fixed camera with an elevation of -25° and an azimuth of 145°.

\subsubsection{Implementation Details}

Octo-small~\cite{2024arXiv240512213O}, a compact high-performance VLA model, is adopted as the backbone for this study. The policy (action) head and value (critic) head are implemented as single fully connected layers. During the post-training stage, the AdamW optimizer is applied with the following hyperparameters: learning rate $10^{-5}$, GAE $\lambda$=$0.95$, discount value factor $\gamma$ = $0.99$, PPO clipping $\epsilon$ = $0.2$, value loss weight = $0.5$, entropy loss weight = $0.0$, action horizon $h$ = $4$, $T_{\text{warmup}}$ = $40k$ steps, batch size = $16$, total training $500k$ steps. Unless otherwise noted, methods use 10 trajectories per task; SFT$_{100}$ uses 100 trajectories per task to emulate data-limited versus data-rich regimes.

\subsubsection{Evaluation Metrics}

Each task is evaluated over 128 episodes under a shared random seed for all models. Reported metrics include: average success rate (Acc.) over the 128 episodes; the 10th percentile of the trajectory-length distribution (Len.); and the mean length of the shortest 10\% of trajectories (Avg(l).) across all episodes.

\subsection{Results Analysis}

Table~\ref{tb1: results} reports results for supervised fine-tuning with 10 demonstrations (SFT), with 100 demonstrations (SFT$_{100}$), standard PPO (no additional demonstrations), and the proposed method (10 demonstrations for initializing the buffer). The proposed approach attains the best average performance across tasks and ranks first on most per-task metrics. Standard PPO post-training of the VLA is highly unstable: performance is comparable only on Reach, while many tasks achieve low accuracy and roughly half exhibit collapse, underscoring the stabilizing effect of the supervised behavior clone within the PPO framework. Compared with SFT (10 demonstrations), the proposed method yields a substantial accuracy gain (0.93 vs. 0.70). Even relative to SFT$_{100}$—using ten times more demonstrations than the seed set used to initialize the demonstration buffer—the proposed method achieves a higher average accuracy (0.93 vs. 0.89). In addition, trajectories under the proposed RL post-training are shorter than those from SFT, indicating that RL may discover strategies superior to the initial demonstrations.

\begin{figure}[!tb]
    \centering
      \includegraphics[width=3.5in]{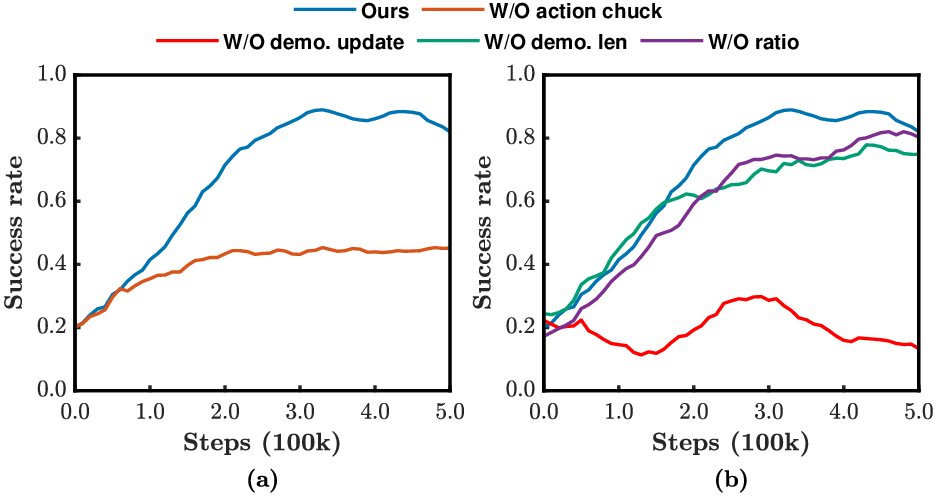}
        \caption{Smoothed performance curve of the ablation study on the MetaWorld Push task. (a) Effect of action chunk in PPO. (b) Effect of the demonstration buffer.} 
     \centering
     \label{fig:ablation}
 \end{figure}

 \begin{figure*}[!tb]
    \centering
      \includegraphics[width=7in]{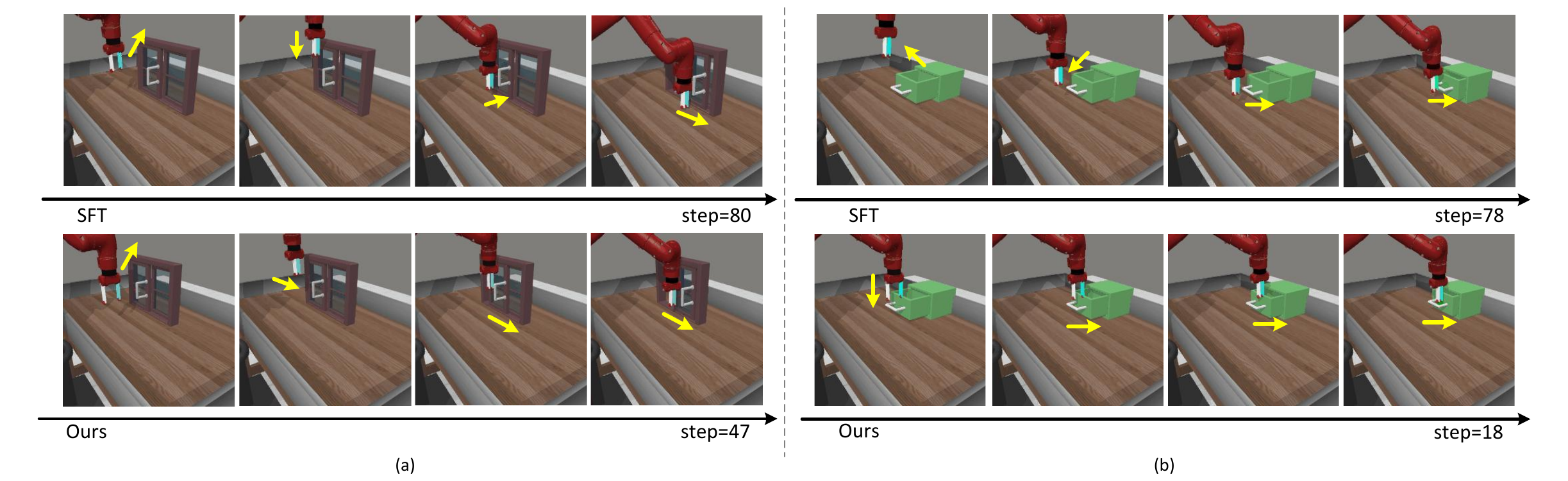}
        \caption{Case study comparing supervised fine-tuning (10 demonstrations) and the proposed method, (a) window open, (b) drawer close} 
     \centering
     \label{fig:case}
 \end{figure*}
 
\subsection{Ablation Study}

The ablation study evaluates two key contributions: the action-chunk design in PPO and the dynamic demonstration buffer for self behavior cloning. As shown in Fig.~\ref{fig:ablation} (a), excluding the action chunk results in a significant performance drop, with the success rate falling to approximately 0.4, compared to 0.77 with the action-chunk design. Despite outperforming supervised fine-tuning (0.34), this reduction highlights the importance of action chunking in enhancing PPO’s effectiveness for post-training VLA models. The results in Fig.~\ref{fig:ablation} (b) underscore the role of the self behavior cloning with demonstration buffer. When no updates are made to the buffer, post-training offers only marginal improvement. Furthermore, if self-generated manipulation trials are added without restrictions, the performance decreases by roughly $10\%$ compared to the performance of the proposed model. These results emphasize the importance of high-quality demonstrations. Additionally, when the PPO and supervised behavior-cloning losses are fixed at a 1:1 ratio, convergence is slower, despite eventual performance reaching similar levels.

\section{Discussion}

Table~\ref{tb1: results} reveals an interesting finding: the number of steps required to complete a given task is reduced compared to SFT (42.17 vs. 66.62). This improvement is likely due to the more efficient strategies for model learning that are not present in the demonstration set. A case study in Fig.~\ref{fig:case} illustrates this: for the ``window open" task, the SFT policy first moves the robotic arm upward, then downward and forward to push the window open, while the proposed policy directly guides the robot to the window handle, reducing the number of steps by nearly half. A similar phenomenon is observed in the ``drawer close" task, where the proposed policy directly controls the robot to close the drawer, rather than first locating the drawer handle as in SFT. These results highlight the benefits of RL-based post-training, which enables the model to discover more efficient task strategies.

\section{Conclusion}

This work introduces an action-chunked PPO algorithm with self-generated demonstrations for post-training VLA models. Consecutive actions are aggregated into action chunks, increasing the effective density of informative feedback. A dynamically updated demonstration buffer continually collects high-quality trajectories generated during training, on which an auxiliary self behavior cloning loss is designed to optimize the policy updates. The relative weight between the PPO objective and the auxiliary loss is adapted online to enhance the stability of the training. Results indicate that RL is a promising paradigm for VLA post-training and provides a practical scheme for continual training. It also facilitates downstream deployment of VLA systems. Future work will focus on real-world applications, large-scale validation, and theoretical analysis.

{\small
\bibliographystyle{IEEEtran}
\bibliography{mybib}
}

\end{document}